\documentclass[conference]{IEEEtran}
\IEEEoverridecommandlockouts

\usepackage{cite}
\usepackage{amsmath,amssymb,amsfonts}
\usepackage{algorithmic}
\usepackage{graphicx}
\usepackage{textcomp}
\usepackage{xcolor}
\usepackage{multirow}
\usepackage{hyperref}
\usepackage{ulem}
\def\BibTeX{{\rm B\kern-.05em{\sc i\kern-.025em b}\kern-.08em
    T\kern-.1667em\lower.7ex\hbox{E}\kern-.125emX}}
\begin{document}

\def\x{{\mathbf x}}
\def\L{{\cal L}}
\normalem
\normalem

\title{PEPL: Precision-Enhanced Pseudo-Labeling for Fine-Grained Image Classification in Semi-Supervised Learning}
%

\author{\IEEEauthorblockN{Bowen Tian$^*$}\thanks{$*$ The first two authors contributed equally to this work.}
\IEEEauthorblockA{\textit{HKUST(GZ)$^\S$}\thanks{$\S$ The Hong Kong University of Science and Technology (Guangzhou)} \\
\textit{$DI^2 Lab^\backprime$}\thanks{$\backprime$ Deep Interdisciplinary Intelligence Lab}\\
Guangzhou, China \\
bowentian@hkust-gz.edu.cn}
\and
\IEEEauthorblockN{Songning Lai$^*$}
\IEEEauthorblockA{\textit{HKUST(GZ)} \\
\textit{$DI^2 Lab$}\\
Guangzhou, China \\
songninglai@hkust-gz.edu.cn}
\and
\IEEEauthorblockN{Lujundong Li}
\IEEEauthorblockA{\textit{HKUST(GZ)} \\
Guangzhou, China \\
1120201534@bit.edu.cn}
\and
\IEEEauthorblockN{Zhihao Shuai}
\IEEEauthorblockA{\textit{HKUST(GZ)} \\
Guangzhou, China \\
zhihaoshuai@hkust-gz.edu.cn}
\and
\IEEEauthorblockN{Runwei Guan}
\IEEEauthorblockA{\textit{HKUST(GZ)} \\
\textit{Institute of Deep Perception Technology, JITRI}\\
\textit{University of Liverpool}\\
Guangzhou, China \\
runwei.guan@liverpool.ac.uk}
\and
\IEEEauthorblockN{Tian Wu}
\IEEEauthorblockA{\textit{Nanchang University} \\
Nanchang, China \\
wutian@ncu.edu.cn}
\and
\IEEEauthorblockN{Yutao Yue$^\dagger$}\thanks{$\dagger$ Correspondence to Yutao Yue \{yutaoyue@hkust-gz.edu.cn\}}\thanks{This work was supported by Guangzhou-HKUST(GZ) Joint Funding Program(Grant No.2023A03J0008), Education Bureau of Guangzhou Municipality}
\IEEEauthorblockA{\textit{HKUST(GZ)} \\
\textit{Institute of Deep Perception Technology, JITRI}\\
\textit{$DI^2 Lab$}\\
Guangzhou, China \\
yutaoyue@hkust-gz.edu.cn}
}

\maketitle


\begin{abstract}

Fine-grained image classification has witnessed significant advancements with the advent of deep learning and computer vision technologies. However, the scarcity of detailed annotations remains a major challenge, especially in scenarios where obtaining high-quality labeled data is costly or time-consuming. To address this limitation, we introduce \underline{\textbf{P}}recision-\underline{\textbf{E}}nhanced \underline{\textbf{P}}seudo-\underline{\textbf{L}}abeling \underline{\textbf{(PEPL)}} approach specifically designed for fine-grained image classification within a semi-supervised learning framework. Our method leverages the abundance of unlabeled data by generating high-quality pseudo-labels that are progressively refined through two key phases: initial pseudo-label generation and semantic-mixed pseudo-label generation. These phases utilize Class Activation Maps (CAMs) to accurately estimate the semantic content and generate refined labels that capture the essential details necessary for fine-grained classification. By focusing on semantic-level information, our approach effectively addresses the limitations of standard data augmentation and image-mixing techniques in preserving critical fine-grained features. We achieve state-of-the-art performance on benchmark datasets, demonstrating significant improvements over existing semi-supervised strategies, with notable boosts in accuracy and robustness.

\end{abstract}


\vspace{-6pt}
\section{Introduction}
\label{sec:intro}
\vspace{-2pt}


Fine-grained image classification \cite{wang2019survey,rong2021human,zhuang2020learning}, which involves distinguishing between visually similar classes, plays a crucial role in various applications such as species identification, product categorization, and medical diagnostics. Despite the remarkable success of deep learning in computer vision \cite{abdullahi2024systematic,rayed2024deep,lai2023multimodal, lai2024learning, tian2024wolf2pack, wu2024maintaining}, achieving high accuracy in fine-grained classification remains challenging due to the scarcity of labeled data and the subtlety of distinguishing features \cite{guo2024fine}.



The limited availability of labeled data, particularly in fine-grained domains, hinders the development of robust models. To mitigate this issue, semi-supervised learning (SSL) \cite{grandvalet2004semi, yang2022class,zeng2023self, 10769516, luo2024contextuality} techniques have been proposed to leverage large amounts of unlabeled data alongside a small labeled dataset. SSL methods, including pseudo-labeling \cite{lee2013pseudo} and consistency regularization \cite{laine2016temporal}, have shown promise in improving model performance with limited supervision. However, existing SSL approaches face significant challenges when applied to fine-grained image classification. Standard data augmentation techniques \cite{cubuk2020randaugment} \cite{cubuk2019autoaugmentlearningaugmentationpolicies} can disrupt critical visual cues, and fine-grained image features are destroyed. Image region mixing may also
overlook the fine details essential for accurate classification \cite{su2021realistic}.


\begin{figure}[t]
\centering
  \includegraphics[width=0.9\columnwidth]{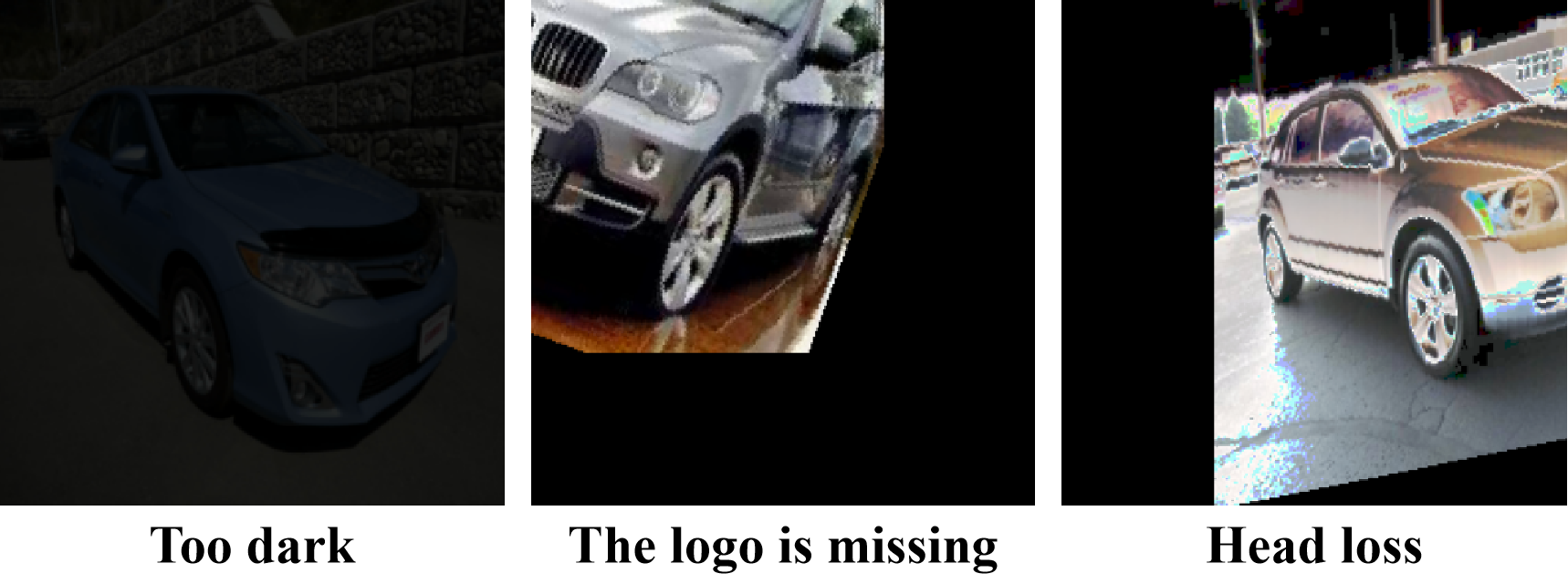}
  \vspace{-10pt}
  \caption{Instances where fine-grained details are corrupted by data augmentation}
  \label{fig:overview}
  \vspace{-20pt}
\end{figure}

To address these challenges, we present a novel \underline{\textbf{P}}recision-\underline{\textbf{E}}nhanced \underline{\textbf{P}}seudo-\underline{\textbf{L}}abeling (PEPL) approach tailored for fine-grained image classification. PEPL leverages CAMs \cite{jiang2021layercam} to generate high-quality pseudo-labels that capture the essential details necessary for fine-grained classification. Specifically, our method consists of two key phases: Initial Pseudo-Label Generation and Semantic-Mixed Pseudo-Label Generation. These phases utilize CAMs\cite{jiang2021layercam,muhammad2020eigen,zhang2024opti} to accurately estimate the semantic content\cite{chen2022class} and generate refined labels that capture the essential details necessary for fine-grained classification. By focusing on semantic-level information, our approach effectively addresses the limitations of standard data augmentation and image-mixing techniques in preserving critical fine-grained features, we have conducted extensive experiments on two commonly used datasets for fine-grained classification, and the results show that our method far exceeds the most advanced and representative semi-supervised methods\cite{yang2022survey,ouali2020overview} at present, with a 13\% improvement in accuracy over the fully supervised model on the CUB\_200\_2011 dataset using 20\% labeled data, and similar results to supervised learning using only 30\% labeled data. 



\begin{figure*}[htbp]
\centering
  \includegraphics[width=0.7\linewidth]{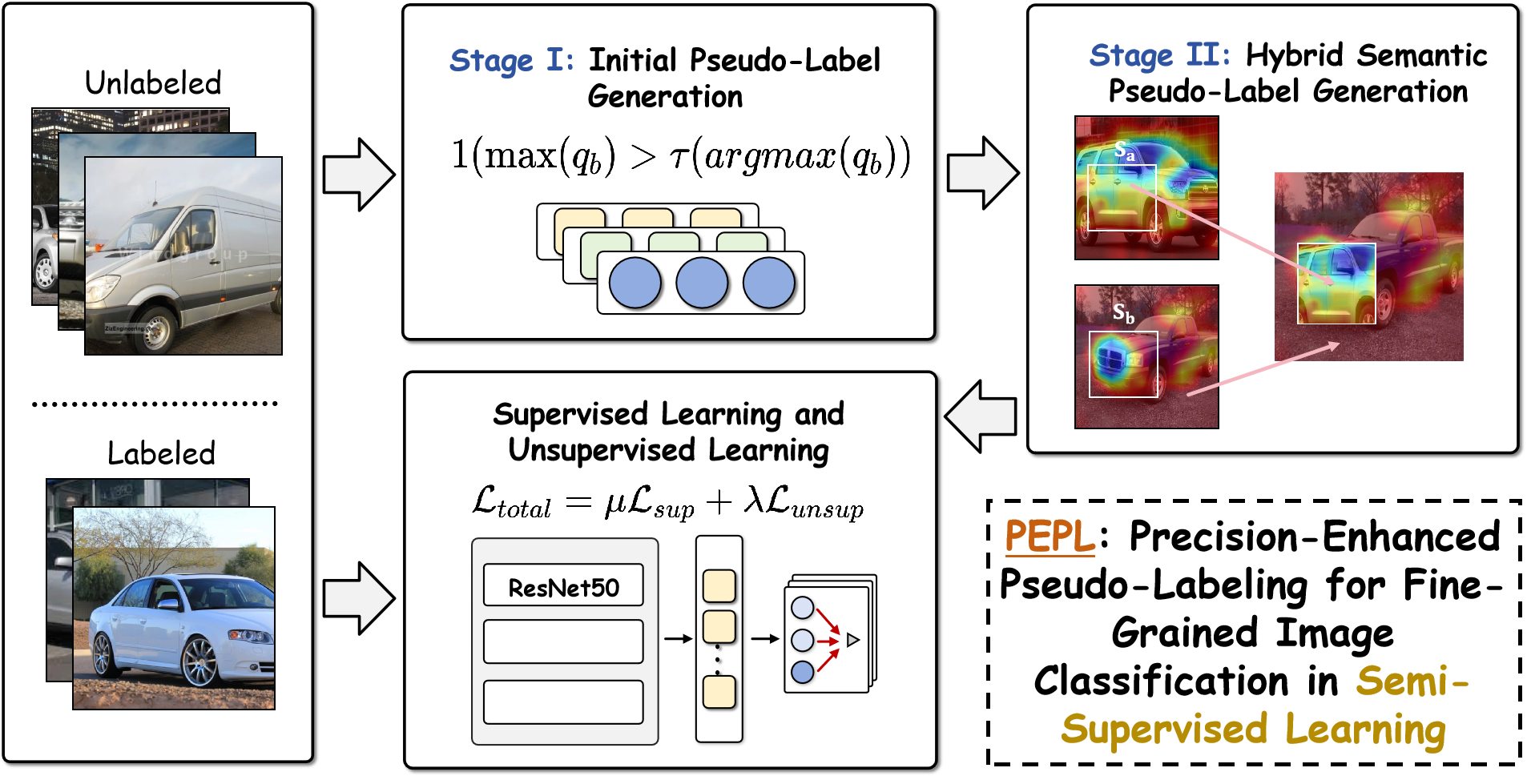}
  \vspace{-10pt}
  \caption{The overview of our proposed methodology.}
  \vspace{-15pt}
  \label{fig:overview}
\end{figure*}


The key contributions of our work are as follows:




(i) We propose the Precision-Enhanced Pseudo-Labeling (PEPL) approach specifically designed for fine-grained image classification.

(ii) Our method generates high-quality pseudo-labels using CAMs, which are progressively refined to enhance the precision of the pseudo-labels.

(iii) We demonstrate significant improvements in performance on benchmark datasets, outperforming existing semi-supervised strategies and achieving state-of-the-art accuracy.

\section{Methods}
\label{sec:met}

\subsection{Stage I: Initial Pseudo-Label Generation}

Inspired by the concept of FreeMatch \cite{wang2022freematch}, our approach relies on the adaptive selection of confidence thresholds, which are dynamically adjusted based on the model's predictive performance on unlabelled data. Rather than adopting a static approach, we holistically evaluate the model's predictions across all classes for each iteration.

After each round of predictions, we apply the following equation to the class outputs of every unlabelled sample. This collective consideration of all classes ensures that the thresholds are not only category-specific but also responsive to the evolving model performance:

First, we holistically consider all categories to determine the overall predictive values for the current iteration. After each round of predictions, we apply the following equation to the class outputs of every unlabelled sample:

\vspace{-8pt}
\begin{equation*}
\hat{\tau}_t = 
\begin{cases}
\frac{1}{C}, & \text{if } t = 0, \\
\beta\tau_{t-1}+(1-\beta)\frac{1}{\mu B}\sum_{b=1}^{\mu B}\max(q_b), & \text{otherwise}
\end{cases}
\end{equation*}
\vspace{-8pt}

\noindent where $C$ represents the total number of categories, $\beta$ is a pre-set hyperparameter that controls the ratio of the EMA, $\mu B$ indicates the batch size of the current unlabeled data, where $B$ indicates the batch size of the labeled data, and $\mu$ is the preset multiple factor, $q_b$ represents the output of the model's predictions, and $\tau_t$ represents the global threshold at step $t$.

To address the issue of class imbalance in the model's predictive capability, we compute an individual model prediction threshold for each class using the following formula:

\vspace{-8pt}
\begin{equation*}
\hat{P}_t(c) = 
\begin{cases}
\frac{1}{C}, & \text{if } t = 0, \\
\beta\hat{P}_{t-1}(c)+(1-\beta)\frac{1}{\mu B}\sum_{b=1}^{\mu B}q_b(c), & \text{otherwise}
\end{cases}
\end{equation*}
\vspace{-8pt}

\noindent where $c$ represents the current category number. 


After obtaining the individual prediction thresholds for each class, we combine the overall threshold and the class-specific thresholds to determine the confidence selection threshold for each class at the current moment:

\vspace{-12pt}
\begin{equation*}
\tau_t(c) = MaxNorm(\hat{p}_t(c))\cdot\tau_t = \frac{\hat{p}_t(c)}{\max{\hat{p}_t(c):c\in[C]}} \cdot \tau_t
\end{equation*}
\vspace{-8pt}

\noindent where $\hat{p}_t(c)$ indicates the prediction threshold of class $c$ at the current time step. This integration of both global and class-wise thresholds allows us to strike a balance between the general performance of the model and the unique characteristics of each class. By doing so, we can effectively select confident predictions for each class, enhancing the reliability of the pseudo-labels in the semi-supervised learning process.

With the thresholds calculated for each class, we can now employ them in the initial generation of pseudo-labels:

\vspace{-8pt}
\begin{equation*}
    \mathbb{I}(\max(q_b) > \tau_t(argmax(q_b))
\end{equation*}
\vspace{-12pt}

\noindent where $\mathbb{I}(\cdot)$ represents the indicator function, which is 1 when the condition is met and 0 otherwise. This step assigns provisional labels to unlabelled samples based on their highest predicted probabilities using derived thresholds. This creates a set of pseudo-labels reflecting the model's confidence, which will be refined in subsequent training iterations of the semi-supervised learning algorithm.

\subsection{Stage II: Hybrid Semantic Pseudo-Label Generation}


The utility of pseudo-labels alone in enhancing model performance is somewhat limited. To better exploit the potential of unlabeled images, we propose a two-stage approach. In Stage I, we randomly blend images and then estimate the semantic information contained in the mixed images. Based on the pseudo-labels generated in the previous step, we create hybrid semantic pseudo-labels for the mixed images. To quantify the semantic composition of the mixed images, we need to measure the semantic correlation between each original image's pixels and their corresponding labels. An effective approach to achieve this is through Class Activation Maps (CAMs), which reveal how regions relate to semantic classes. We initially employ an attention mechanism \cite{wang2017residual} to compute the class activation map for the input image. Let the feature map at the $l$th layer be represented as:

\vspace{-10pt}
\begin{equation*}
    F_l(\hat{I}_i) \in \mathbf{R}^{c\times h \times w}
\end{equation*}
\vspace{-16pt}

\noindent where $ I_i$ represents the input image, and $c,h,w$ represents the number of categories, height and width of the feature map, respectively. We can match the activation map from the $l$ th layer to the size of the input image using upsampling operations:

\vspace{-15pt}
\begin{equation*}
    CAM(I_i) = \phi(\sum_{t=0}^d w_{y_i}^{l}F_l(I_i))
\end{equation*}
\vspace{-12pt}

\noindent where $CAM(\cdot)$ represents an activation map of the same size as the input image, and $\phi(\cdot)$ represents the upsampling operation. Next, we normalize the $CAM(I_i)$ to obtain a map with a sum equal to 1:

\vspace{-8pt}
\begin{equation*}
    S(I_i)=\frac{CAM(I_i)}{sum(CAM(I_i))}
\end{equation*}
\vspace{-8pt}

\noindent where $S(\cdot)$ represents the activation map that has been normalized. This step involves transforming the CAM associated with the image $I_i$ such that the sum of all its values equals unity. When blending images, we infer the label of the mixed image based on the semantic proportions of each component in the original images:

\vspace{-13pt}
\begin{equation*}
    \rho_a = 1 - sum(M_{\lambda^a}\odot S(I_a)),
\end{equation*}
\vspace{-15pt}

\vspace{-15pt}
\begin{equation*}
        \rho_b = sum(M_{\lambda^b}\odot S(I_b))
\end{equation*}
\vspace{-16pt}




The process of estimating the label for the blended image involves considering the relative semantic contributions of the individual parts from the original images. For each blended image, we estimate the proportions $\rho_a$ and $\rho_b$ of the semantic pseudo-labels $a$ and $b$ respectively. $M_{\lambda^a}$ represents the part of input $a$ that is removed, and $M_{\lambda^b}$ represents the part of input $b$ that is blended into input $a$. We derive $\rho_a$ by subtracting the removed portion from 1, and $\rho_b$ by estimating the semantic proportion of the blended part. These proportions reflect the combined semantic content of the blended image.

For each batch of unlabeled data, we first generate preliminary pseudo-labels. Then, we randomly combine these pseudo-labelled samples to create mixed instances along with their corresponding hybrid semantic pseudo-labels. These hybrid labels are used to iteratively refine and optimize the model during training.


\subsection{Loss Function for Whole Framework}
We can divide the overall loss function into supervised loss $\mathcal{L}_{sup}$ and unsupervised loss $\mathcal{L}_{unsup}$. The calculation of the loss function can be expressed as follows:

\vspace{-12pt}
\begin{equation*}
    \mathcal{L}_{sup} = \mathcal{H}(p_m(x_i|\theta),y_i),
\end{equation*}
\vspace{-20pt}

\vspace{-15pt}
\begin{equation*}
    \mathcal{L}_{unsup} = \mathcal{H}(p_m(x_a|\theta),y_a) \cdot \rho_a + \mathcal{H}(p_m(x_b|\theta),y_b) \cdot \rho_b,
\end{equation*}
\vspace{-15pt}

\vspace{-12pt}
\begin{equation*}
    \mathcal{L}_{total} = \gamma \mathcal{L}_{sup} + \lambda \mathcal{L}_{unsup}
\end{equation*}
\vspace{-15pt}

\noindent where $p_m$ represents the predicted output for input $x_i$ when the parameter is $\theta$, and $\mathcal{H}(\cdot) $ represents the cross-entropy loss function. $y_a$ and $y_b$ represent two semantic pseudo-labels generated by the steps above. $\gamma$ and $\lambda$ represent the weights of supervised and unsupervised losses, respectively.

\section{Experiment and Result}
\label{sec:exp}

\begin{table*}[htbp]
\vspace{-10pt}
\centering
\caption{Comparison with other semi-supervised methods}
\label{tab:cub}
\begin{tabular}{ccccccc}
\hline
Dataset & Method & 10\%Label\ $\uparrow$ & 20\%Label\ $\uparrow$ & 30\%Label\ $\uparrow$ & 100\%Label\ $\uparrow$\\
\hline
\multirow{6}*{CUB\_200\_2011} & Supervised-Only & 28.61 & 51.87 & 65.77 & 85.76 \\
~ & Pi-Model & 25.52 & 50.65 & 60.79 & 75.56 \\
~ & Pseudo-Label & 32.71 & 54.42 & 68.93 & 86.77 \\
~ & FlexMatch & 30.61 & 55.71 & 70.15 & 87.98 \\
~ & FreeMatch & 30.78 & 56.68 & 67.62 & 88.39 \\
~ & \textbf{PEPL(Ours)} & \textbf{38.53} & \textbf{64.60} & \textbf{76.97} & \textbf{88.75} \\
\hline
\multirow{6}*{Stanford Car} & Supervised-Only & 24.54 & 54.13 & 70.71 & 90.09 \\
~ & Pi-Model & 13.07 & 48.52 & 67.75 & 85.49 \\
~ & Pseudo-Label & 26.12 & 60.10 & 74.35 & 90.19 \\
~ & FlexMatch & 26.70 & 61.32 & 73.31 & 90.79 \\
~ & FreeMatch & 26.10 & 62.67 & 75.97 & 89.26 \\
~ & \textbf{PEPL(Ours)} & \textbf{32.72} & \textbf{74.79} & \textbf{86.52} & \textbf{91.09} \\
\hline
\vspace{-15pt}
\end{tabular}
\end{table*}


\subsection{Setup}

\textbf{Datasets. }To evaluate the effectiveness of PEPL, we conducted experiments on two standard fine-grained classification datasets: CUB\_200\_2011 \cite{WahCUB_200_2011} and Stanford Cars \cite{krause20133d}. The first dataset, introduced by Caltech in 2010, comprises 11,788 images across 200 bird species, with 5,994 images for training and 5,794 for testing. This dataset is widely used as a benchmark for fine-grained classification and recognition research. Stanford Cars, released by Stanford AI Lab in 2013, includes 16,185 images of 196 car models, with 8,144 images for training and 8,041 for testing. The dataset is designed for fine-grained classification tasks and categorizes cars by brand, model, and year. 

\begin{table}
\centering
\vspace{-10pt}
\caption{An exploration of the availability of semantic information}
\label{tab:ablation}
\begin{tabular}{ccc}
\hline
Label Ratio & Semantic Aware & without Semantic Aware \\
\hline
10\% & 38.53 & 27.47 \\
20\% & 64.60 & 56.23 \\
30\% & 76.97 & 72.50 \\
100\% & 88.75 & 85.08 \\
\hline
\vspace{-25pt}
\end{tabular}
\end{table}

\textbf{Settings. }We conducted experiments using a single NVIDIA A800 80G GPU. The pre-trained ResNet50 on ImageNet served as the base classification model. The overall training was set for 200 epochs, with a batch size of 16 for labeled data. For unlabeled data, the batch size was set to 112 ($\mu = 7$). The initial learning rate was 0.01, decreasing by a factor of 0.1 every 80 epochs. After reaching 0.0001, a cosine annealing scheduler was applied to gradually reduce the learning rate to 0 over the last 40 epochs. The hyperparameter $\beta$ for pseudo-label generation in Stage I was set to 0.999 to ensure a stable growth trend. Both the loss weights ($\gamma$ and $\lambda$) for supervised and unsupervised learning were set to 1.

\subsection{Results and Analysis}

\textbf{Evaluation Metric. }We chose multi-classification accuracy as our evaluation metric, defined below:
$$Accuracy = \frac{TP+TN}{ALL}$$
where TP+TN represents the number of samples with all the correct classifications, and ALL represents the total number of samples.

\textbf{Performance. }The main experimental results are summarized in Table \ref{tab:cub}. We compared our method with classic semi-supervised learning approaches, Pi-Model \cite{laine2016temporal} and Pseudo-Label \cite{article}, as well as sota methods, FlexMatch \cite{zhang2021flexmatch} and FreeMatch \cite{wang2022freematch}, under scenarios with 10\%, 20\%, and 30\% of the total data labeled, and also when all label data were used. We also compared with purely supervised learning (Supervised-Only). The perturbation method of Pi-Model and the strong augmentation methods of FlexMatch and FreeMatch all used RandAugment \cite{cubuk2020randaugment} technology. The classification accuracy on the two datasets clearly demonstrates that our proposed PEPL method consistently outperforms other semi-supervised learning methods under different label proportions. Using just 30\% of the labels, our method achieves comparable results to supervised training with 100\% of the label data. With 10\% and 20\% of the label data, our method outperforms state-of-the-art semi-supervised methods by approximately 8\%, and improves accuracy by about 10\% to 13\% compared to purely supervised training. These results fully demonstrate the effectiveness of our proposed PEPL semi-supervised learning framework in enhancing fine-grained classification performance across different datasets.

\noindent \textbf{Ablation Study. }To further validate the effectiveness of semantically mixed pseudo-labels introduced by PEPL, we compared it with the method of directly mixing and generating pseudo-labels without semantic mixing on the CUB. As shown in Table \ref{tab:ablation} and combined with Table \ref{tab:cub}, we find that while direct mixing without semantic mixing still achieves some improvement compared to purely supervised learning, adding semantic mixing results in an additional performance gain of about 4\% to 9\%. This fully demonstrates the rationale behind introducing semantically mixed pseudo-labels in PEPL.

\begin{figure}
    \centering
    \includegraphics[width=0.75\linewidth]{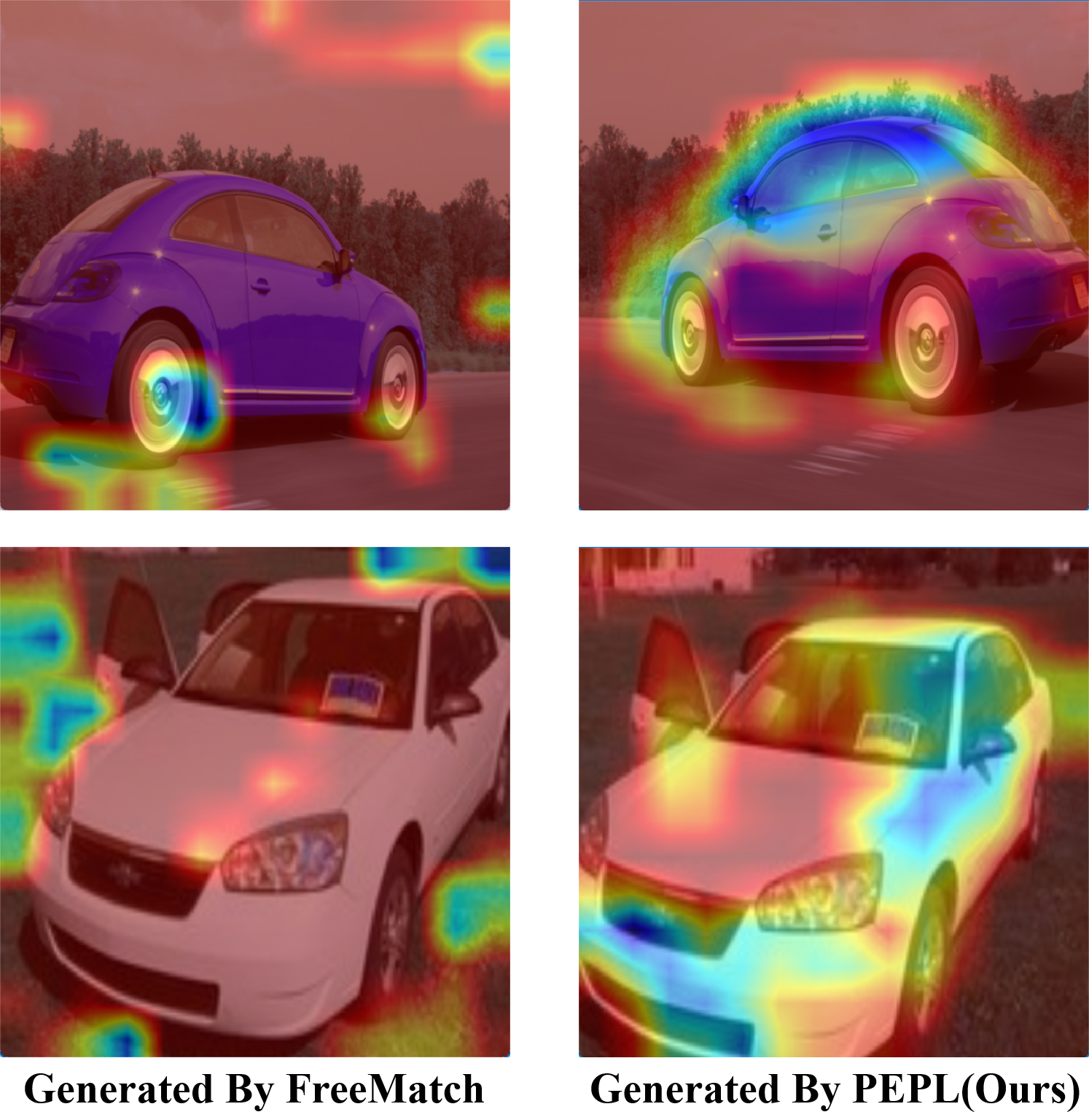}
    \vspace{-9pt}
    \caption{Compared with method FreeMatch, the classifier obtained by method PEPL focuses more on fine-grained features such as car logos, lights, mirrors, and other more distinguishing parts.}
    \label{fig:4part}
    \vspace{-12pt}
\end{figure}

\noindent \textbf{Case Study. }To more intuitively demonstrate the superiority of the PEPL method, we exported models trained using the FreeMatch method and the PEPL method for semi-supervised training on 30\% labeled data. We calculated class attention maps using the output of the last convolutional layer and visualized them. As shown in Figure \ref{fig:4part}, it is evident that the class attention maps based on the PEPL method focus more on areas where the current class may have fine-grained differences with other classes (such as car logos and rearview mirrors). This intuitively indicates that the PEPL method can better enhance the model's perception of fine-grained features.

\section{Conclusion}
\label{sec:con}


In this paper, we introduced the PEPL method, which effectively addresses the challenges faced by semi-supervised learning methods in the domain of fine-grained image classification. By leveraging CAMs to generate high-quality pseudo-labels, PEPL overcomes the limitations of standard data augmentation and image-mixing techniques. The simplicity and effectiveness of PEPL make it a valuable addition to the toolkit of researchers and practitioners working in fine-grained classification, alleviating the exceptionally severe label scarcity problem. Its flexibility and strong performance position PEPL as a method that can significantly advance state of the art in semi-supervised learning and inspire further research into innovative approaches for fine-grained image classification.

\bibliographystyle{IEEEbib}
\bibliography{strings}

\end{document}